\documentclass[a4paper]{article}

\usepackage{a4wide}
\usepackage{fancyhdr}
\usepackage{comment}
\usepackage{amsmath,amssymb}
\usepackage{bm}
\usepackage{graphicx}
\usepackage[dvipdfmx]{color}
\usepackage{ascmac}
\usepackage{url}
\usepackage{siunitx}
\usepackage{float}
\usepackage{slashbox}
\usepackage{here}
\usepackage{txfonts}
\usepackage{listings}
\usepackage{tikz}

\newcolumntype{G}{>{\columncolor[gray]{0.8}}c}
\newcommand{\hcircle}[1]{\mathrm{\scriptsize \textcircled{\tiny \raisebox{-0.5pt}[0ex][0ex]{H}}}}

\title{Hierarchical Model Selection for Graph Neural Netoworks} 
\author{Yuga Oishi\footnotemark[1] \;\;\;\;\;Ken Kaneiwa\footnotemark[1]}
\date{\footnotesize	\footnotemark[1] Department of Computer and Network Engineering, Graduate School of Informatics and Engineering, \\The University of Electro-Communications, Tokyo, Japan}

\begin{document}
\maketitle

\begin{abstract}\noindent
Node classification on graph data is a major problem, and various graph neural networks (GNNs) have been proposed. 
Variants of GNNs such as H2GCN and CPF outperform graph convolutional networks (GCNs) by improving on the weaknesses of the traditional GNN. 
However, there are some graph data which these GNN variants fail to perform well than other GNNs in the node classification task.
This is because H2GCN has a feature thinning on graph data with high average degree, and CPF gives rise to a problem about label-propagation suitability. 
Accordingly, we propose a hierarchical model selection framework (HMSF) that selects an appropriate GNN model by analyzing the indicators of each graph data.
In the experiment, we show that the model selected by our HMSF achieves high performance on node classification for various types of graph data.
\end{abstract}

\section{Introduction}
Graph data can represent various networks in the real world, such as citation networks of papers and link relations of Web pages. 
Node classification on graph data is one of the significant tasks in machine learning, and various models \cite{node2vec, lp, minami} have been proposed to solve this. 
Among these models, GCN (graph convolutional network) \cite{gcn} has attract a lot of attention by outperforming traditional methods.
Furthermore, various GNNs (graph neural networks) \cite{gnn, gat, gin, jknet} based on GCN have been proposed, and achieve state-of-the-art performances.\\

\noindent
There are concepts called homophily and heterophily which indicate the distribution of labels on graph data.
Most GNNs assume graph data with homophily that nodes with the same label exist nearby. 
So that they may not perform well on graph data with heterophily that nodes with different labels are nearby.
To address this problem, H2GCN \cite{h2} aggregates and combines the features of neighboring nodes by depths.
As a knowledge distillation framework using GNN,  CPF (Combination of Parameterized label propagation and Feature transformation \cite{cpf}) trains student model based on teacher GNN model.
In CPF, the student model better performs than teacher model by fully leveraging information of graph data.
However, the high expressive power of H2GCN and CPF may hurt learning on some data, and it is difficult for a particular GNN to outperform other GNNs on all graph data.\\

\noindent
Quantitative indicators are required to analyze the properties of graph data. 
Average degree is an indicator of the structural properties of graph data which represents the average number of edges a node has.
In addition, edge homophily ratio is an indicator about homophily and heterophily on graph data which indicates the ratio of edges connecting nodes with the same label.\\

\noindent
In this paper, we propose a hierarchical model selection framework (HMSF) which selects an appropriate GNN model by analyzing the indicators of graph data. 
First, we analyze how the average degree affects the training of GCN and H2GCN. 
Based on this analysis, we demonstrate a method to select a GNN model trained as a teacher model.
Then, edge homophily ratio is estimated by predicting the labels of all nodes using the output of this teacher model. 
Based on this edge homophily ratio, we analyze the effectiveness of the student model in CPF and demonstrate a method to decide whether HMSF selects CPF.\\
\\\noindent
The contributions of this paper are summarized as follows:
\begin{itemize}
\item \textbf{Weaknesses of Previous Works}: We point out the weaknesses of previous works that performs poorly on several datasets. We demonstrate that H2GCN has a feature thinning on graph data with high average degree, and CPF gives rise to a problem about label-propagation suitability.
\item \textbf{Model Selection by Average Degree and Edge Homophily Ratio}: We propose a hierarchical model selection framework (HMSF) which can select an appropriate GNN model for each dataset. HMSF uses average degree and edge homophily ratio to determine the usefulness of previous GNN models. 
\item \textbf{Extensive Experiments}: Our evaluation experiments show that HMSF can select model which achieves high performance for nine benchmark datasets. Furthermore, we demonstrate some specific examples that H2GCN and CPF does not perform well based on our analyses: (i) the aggregated features at nodes with high degree are thinned in H2GCN, and (ii) CPF using LP is not suitable for graph data with heterophily.
\end{itemize}
\noindent
The rest of this paper is structured follows: In Section 2, we explain various concepts of graph neural networks. In Section 3, we analyze the limitations of previous studies arising from the properties of graph data. In Section 4, we present indicators about graph and formulate the HMSF using these indicators. In Section 5, we discuss the related works. In Section 6, we conduct experiments and analyze about HMSF across multiple datasets. In Section 7, we conclude the paper and discuss future work.
\section{Notation and Preliminaries}

\subsection{Semi-supervised Learning for Node Classification}
We describe semi-supervised learning for the node classification in graph data. Let $G = (V, E)$ be a graph with a node set $V$ and an edge set $E$. Each node $v \in V_L$ contained in a subset of the node set $V_L \subset V$ has a label $y_v \in Y$ where $Y$ is a label set. The object of node classification is to predict the label of an unlabeled node $v \in V_U = V \backslash V_L$. We can use an $F$-dimensional feature vector $X_v\in \mathbb{R}^F$ for each node $v\in V$ for learning.

\subsection{GCN (Graph Convolutional Network)}
GCN \cite{gcn} is a basic GNN model that aggregates the feature vectors of neighboring nodes and average them. Let $N_{i}(v)$ be the set of nodes that exist exactly $i$ hops away from node $v$. The output $h_{v}^{(k)}$ of node $v$ at the $k$-th layer in GCN is expressed as:

\begin{eqnarray}
{h'}_{v}^{(k)} & = & \sum_{u \in N_{1}(v) \cup \{v\}} h_{u}^{(k-1)} (d_{v,1}+1)^{-1/2}(d_{u,1}+1)^{-1/2};  \nonumber \\
h_{v}^{(k)} &= & \sigma \Bigl( {h'}_{v}^{(k)} W_k \Bigr), \nonumber
\end{eqnarray}
where $h_{v}^{(0)} = X_{v}$, $d_{v,i} = |N_{i}(v)|$, $W_k$ is the learnable weight matrix of the $k$-th layer, and $\sigma$ is the activation function. The output layer of GCN (the $K$-th layer) outputs the class of node $v$ as follows:

\begin{equation}
f_{\mbox{\it gcn}}(v ; W_0,...,W_K) = \mathrm{softmax}(h_v^{(K)}). \nonumber
\end{equation}

\subsection{H2GCN}
H2GCN \cite{h2} is an expressive variant of GNNs that distinguishes the aggregated feature vectors by depth. The aggregated feature vector of nodes at depth $i$ from node $v$ in the $k$-th layer is represented as follows:

\begin{equation}
r_{v,i}^{(k)} = \sum_{u \in N_{i}(v)} r_{u}^{(k-1)} d_{v,i}^{-1/2}d_{u,i}^{-1/2}. \nonumber
\end{equation}
The representations of the 0-th layer $r_v^{(0)}$, the middle layer $r_v^{(k)}\ (1 \leq k \leq K)$, and the terminal layer $r_v^{(\mathrm{final})}$ in H2GCN are given by the following equations, respectively:

\begin{eqnarray}
r_{v}^{(0)} & = & \sigma \bigl( X_{v} W_e \bigr); \nonumber\\
r_{v}^{(k)} & = & \mathrm{CONCAT} \bigl( r_{v,1}^{(k)}, r_{v,2}^{(k)} \bigr); \nonumber\\
r_{v}^{(\mathrm{final})} & = &  \mathrm{CONCAT} \bigl( r_{v}^{(0)}, r_{v}^{(1)}, ..., r_{v}^{(K)} \bigr), \nonumber
\end{eqnarray}
where $W_e\in\mathbb{R}^{F \times p}$ and $W_c \in\mathbb{R}^{(2^{K+1} - 1)p \times |Y|}$ are the learnable weight matrix, and $\rm{CONCAT}$ is the operation to concatenate the vectors. The output layer of H2GCN outputs the class of node $v$ as follows:

\begin{equation}
f_{\mbox{\it h2gcn}} ( v ; W_e,W_c) = \mathrm{softmax} \bigl( r_{v}^{(\mathrm{final})} W_c \bigr). \nonumber 
\end{equation}

\subsection{CPF}
CPF (Combination of Parameterized label propagation and Feature transformation) \cite{cpf} is a knowledge distillation framework that uses a trained GNN as a teacher model and improves accuracy by additional training in the student model. CPF learns feature-vector and graph-structure information, which GNNs do not fully utilize, using MLP (multi-layer perceptron) \cite{mlp} and LP (label propagation) \cite{lp}, respectively. The FT (feature transformation) module for the feature vector $X_v$ of node $v$, and the PLP (parameterized label propagation) module for node $v$ and its neighboring nodes $N_1(v)$ are given as follows:

\begin{eqnarray}
f_{\mbox{\it ft}}(v) & = & \mathrm{softmax(MLP}(X_v)); \nonumber\\
f_{\mbox{\it plp}}^{(k)}(v) & = & \sum_{u \in N_{1}(v) \cup \{v\}} w_{uv} f_{lp}^{(k-1)}(u). \nonumber
\end{eqnarray}
Here, the PLP module is initialized as follows:

\begin{equation}
f_{\mbox{\it plp}}^{(0)}(v) = 
    \begin{cases}
(0,...,1,...,0) \in \mathbb{R}^{|Y|} & \forall v \in V_{L};\\
(\frac{1}{|Y|},...,\frac{1}{|Y|},...,\frac{1}{|Y|}) \in \mathbb{R}^{|Y|} &  \forall v \in V_{U}.   \nonumber
    \end{cases}
\end{equation}
$w_{uv}$ represents the weight for the edge between nodes $u$ and $v$, which is expressed by the following equation using the confidence score $c_{v}\in \mathbb{R}$ assigned to each node $v$.

\begin{equation}
w_{uv} = \frac{exp(c_{u})}{\sum_{u' \in N_{1}(v) \cup \{v\}} exp(c_{u'})} \nonumber
\end{equation}
The output of the CPF combining the two modules is represented by the following equation using the learnable balance parameters $\alpha\in [0, 1]$ for PLP and FT:

\begin{equation}
f_{\mbox{\it cpf}}^{(k)}\ (v ; \hcircle{}) = \alpha_v \sum_{u \in N_{1}(v) \cup \{v\}} w_{uv} f_{\mbox{\it cpf}}^{(k-1)}\ (u) + (1 - \alpha_v) f_{\mbox{\it ft}}(v), \nonumber
\end{equation}
where $\hcircle{}$ is the parameter set used by CPF, and $f_{\mbox{\it cpf}}^{(0)}\ (v)$ is initialized in the same way as $f_{\mbox{\it plp}}^{(0)}(v)$. The final CPF objective function is expressed in the following equation using the output $f_{\mbox{\it gnn}}(v)$ of the teacher model and the L2 norm $||\cdot||_{2}$:

\begin{equation}
\underset{\hcircle{}}{\mathrm{min}} \sum_{u \in V_U} || f_{\mbox{\it gnn}}(v) - f_{\mbox{\it cpf}}^{(K)}\ (v ; \hcircle{}) ||_{2}. \nonumber
\end{equation}

\subsection{Edge Homophily Ratio}
The edge homophily ratio $h$ is an indicator which expresses the ratio of edges in a graph that connect nodes with the same label. Graph data with homophily have a high $h$, whereas graph data with heterophily have a low $h$. The edge homophily ratio is calculated as follows:

\begin{equation}
h = \frac{|\{(u, v) : (u, v) \in E \land y_u = y_v \}|}{|E|}. \nonumber
\end{equation}

\section{Problems in Previous Works}
\subsection{Feature Thinning in H2GCN}
\begin{figure}[H]
\begin{center}
  \includegraphics[scale=0.6]{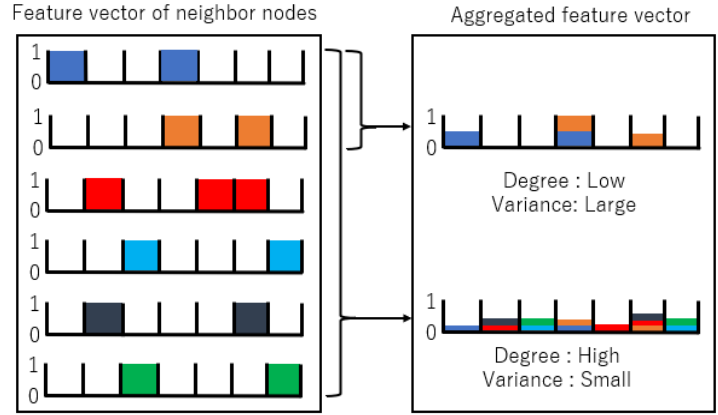}
  \caption{An example of aggregation at nodes with low/high degree}\label{fig:aggr}
\end{center}
\end{figure}

We demonstrate that H2GCN does not perform well on graph data with a high average degree. For simplicity, we consider the zeroth layer $r_v^{(0)}$ in H2GCN as a vector with 0 and 1 (e.g., [0.0,... ,1.0,... ,1.0]). The final layer $r_v^{(\mathrm{final})} = \mathrm{CONCAT} \bigl( r_{v}^{(0)}, r_{v}^{(1)}, ..., r_{v}^{(K)} \bigr)$ is the combined vector representation of each layer, $r_v^{(0)}$, $r_{v}^{(1)} = \mathrm{CONCAT} (r_{v,1}^{(1)}, r_{v,2}^{(1)})$, etc. Here, $r_{v,1}^{(1)} = \sum_{u \in N_{1}(v)} r_{u}^{(0)} d_{v,1}^{-1/2}d_{u,1}^{-1/2}$ is a vector that aggregates the features of the neighbor nodes of $v$, and the feature amount aggregated from each neighbor node $u \in N_{1}(v)$ takes the value $d_{v,1}^{-1/2}d_{u,1}^{-1/2}$ normalized by the degree $d_{v,1} = |N_{1}(v)|$ of node $v$ and the degree $d_{u,1} = |N_{1}(u)|$ of neighbor node $u$. 
Therefore, when the degree of $u$ and $v$ is very high, the feature amount aggregated from $u$ takes very small value. 
Furthermore, when neighboring nodes have different features, $r_{v,1}^{(1)}$ is flattened to a small value owing to the aggregation of various smaller features (e.g., [0.2,... ,0.18,... ,0.21]). 
In this case, the features of the self-node can be distinguished because they are expressed as a vector with 0 and 1 in $r_v^{(0)}$, but the features of the vectors aggregated from neighboring nodes are difficult to distinguish because they take close values. 
This phenomenon also occurs in each $r_{v,i}^{(k)}$ aggregated at other layers and depths. 
Figure \ref{fig:aggr} illustrates the aggregated vectors when the features of neighboring nodes are aggregated at nodes with low/high degree. 

\noindent
The normalization of aggregated features is applied in GCN as well.
However, in GCN, the feature vectors of node and its neighbors are aggregated into one vector and not separated like H2GCN.
So that, even if the aggregated features take very similar values, GCN can learn small differences of features about neighbors.
On the other hand, the large differences of features in zeroth layer constructed 0 and 1 are emphasized in H2GCN.
This leads to H2GCN ignoring small differences of features in aggregated vectors.
In other words, the high expressive power of H2GCN may not perform well when the average degree is high.
For the reason above, GCN is more effective compared to H2GCN for graph data with high average degree.

\subsection{Label Propagation Suitability}
\begin{figure}[H]
\begin{center}
  \includegraphics[scale=0.6]{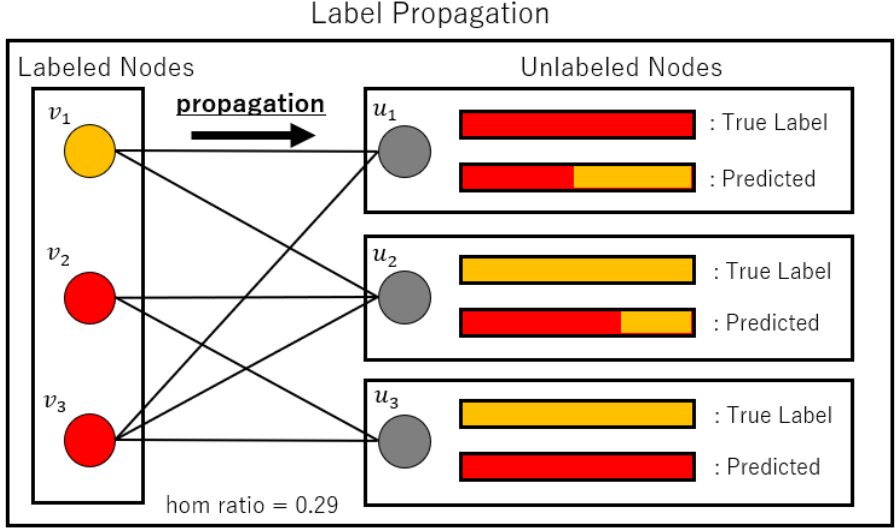}
  \caption{An example of applying LP to a graph data with heterophily (edge homophily ratio = 0.29). The labels of labeled nodes $v_i$ are propagated to unlabeled nodes $u_i$.}\label{fig:lp}
\end{center}
\end{figure}

LP is an algorithm that propagates the label of a node to its neighbors based on the assumption that nodes with same label exist nearby. 
It means that LP is useful for graph data with homophily. 
On the other hand, graph data with heterophily does not satisfy the assumption of LP since most of the edges connect nodes with different labels.
Figure \ref{fig:lp} illustrates an example of applying LP to a graph data with heterophily.

\noindent
In the case of Figure \ref{fig:lp}, the label (red) of labeled nodes $v_2$ and $v_3$ are different from the true label (yellow) of neighboring node $u_2$ and $u_3$. 
Therefore, LP predicts incorrect label (red) to $u_2$ and $u_3$ by propagation of label.
In CPF, the student model uses LP to get structural information of graph data that is not captured in teacher model. 
However, if CPF is applied to graph data with heterophily, CPF are not always helpful owing to the propagation of different labels by LP.
In Section \ref{experiment}, we prove that the student model of CPF actually performs poorly compared to the teacher GNN model on graph data with heterophily.

\section{Hierarchical Model Selection Framework}
\begin{figure}[H]
\centering
  \includegraphics[width=15.0cm]{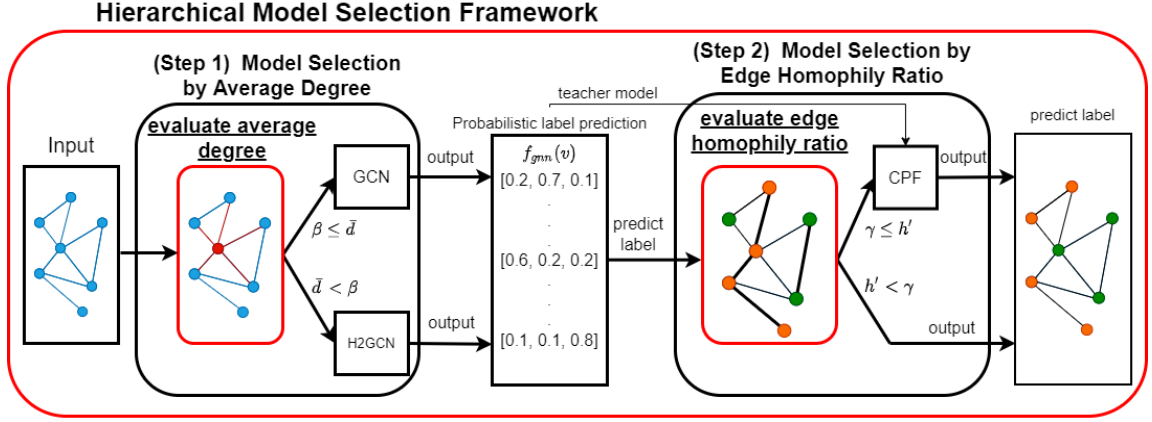}
  \caption{Process Flow in the Hierarchical Model Selection Framework}\label{fig:sequence}
\end{figure}

We propose HMSF based on the analysis in the previous section.  The overall process flow in HMSF is presented in Figure \ref{fig:sequence}. We assume that weights are learned in GCN by minimizing the following cross-entropy loss:
\begin{equation}
\hat{W}_0,...,\hat{W}_{K_1} = \underset{W_0,...,W_{K_1}}{\mathrm{argmin}} \Bigl(-\sum_{v \in V_L}\!f_{\mbox{\it gcn}}(v ; W_0,...,W_{K_1}) \mathrm{ln} (y_v) \Bigr). \nonumber
\end{equation}
In H2GCN, the weights are also learned by minimizing the following cross-entropy loss:
\begin{equation}
\hat{W}_e,\hat{W}_c = \underset{W_e,W_c}{\mathrm{argmin}} \Bigl( - \sum_{v \in V_L} f_{\mbox{\it h2gcn}}(v ; W_e,W_c) \mathrm{ln} (y_v) \Bigr). \nonumber
\end{equation}

\subsection{(Step 1) : Model Selection by Average Degree}

In HMSF, we first evaluate the average degree $\bar{d}$ via the following equation:
\begin{equation}
\bar{d} = \frac{\sum_{v \in V} |N_{1}(v)|}{|V|}. \nonumber
\end{equation}
As in the left-hand side of Figure \ref{fig:sequence}, depending on the average degree, we then determine the trained GNN model as follows:

\begin{equation}
f_{\mbox{\it gnn}}(v) =
    \begin{cases}
f_{\mbox{\it gcn}}(v ; \hat{W}_0,...,\hat{W}_{K_1}) & \mathrm{if}\quad \beta \leq \bar{d}; \\
f_{\mbox{\it h2gcn}}(v ; \hat{W}_e, \hat{W}_c) & \mathrm{otherwise},   \nonumber
    \end{cases}
\end{equation}
where the parameter $\beta$ is a boundary to determine whether the average degree of the graph data is high or not. In this way, we can select GCN to be used for graph data with high average degree; H2GCN is selected otherwise.

\subsection{(Step 2) : Model Selection by Edge Homophily Ratio}
Edge homophily ratio is calculated using the labels of all nodes in the inputted graph data. However, since we can only use the labels of nodes in the training data, the exact edge homophily ratio cannot be calculated. In this framework, we use the output $f_{\mbox{\it gnn}}(v)$ of the trained GNN model to give all nodes the predicted label $\ y'_v = \mathrm{argmax}(f_{\mbox{\it gnn}}(v))$. This allows us to estimate the edge homophily ratio using the predicted labels for all nodes by computing the following equation:

\begin{equation}
\frac{|\{(u, v)\!:\!(u, v)\!\in\!E\!\land\!\mathrm{{\scriptstyle argmax}}(f_{\mbox{\it gnn}}(u))\!=\!\mathrm{{\scriptstyle argmax}}(f_{\mbox{\it gnn}}(v)) \}|}{|E|}. \nonumber
\end{equation}
The CPF parameter $\hcircle {}$ is learned by training the student model $f_{\mbox{\it cpf}}^{K_2}\ (v ; \hcircle{})$ to minimize the error in the output of the teacher model via the following equation:

\begin{equation}
\hat{\hcircle{}} = \underset{\hcircle{}}{\mathrm{argmin}} \Bigl( \sum_{u \in V_U} || f_{\mbox{\it gnn}}(v) - f_{\mbox{\it cpf}}^{K_2}\ (v ; \hcircle{}) ||_{2} \Bigr). \nonumber
\end{equation}
As in the right-hand side of Figure \ref{fig:sequence}, we determine the final output of the framework as follows:

\begin{equation}
f_{\mbox{\it hmsf}}\ (v) =
    \begin{cases}
f_{\mbox{\it cpf}}^{K_2}\ (v ; \hat{\hcircle{}}) & \mathrm{if}\quad \gamma \leq h'; \\
f_{\mbox{\it gnn}}(v) & \mathrm{otherwise},  \nonumber
    \end{cases}
\end{equation}
where h' is the estimated edge homophily ratio and the parameter $\gamma$ is a boundary to determine whether the data has a high edge homophily ratio. Thus, we can select $f_{\mbox{\it gnn}}(v)$ for graph data with low edge homophily ratio; CPF, which takes $f_{\mbox{\it gnn}}(v)$ as the teacher model, is selected otherwise.

\section{Related Work}
We discuss related work on model selection and GNN in this section. 
AutoHEnsGNN\cite{autohensgnn} is one of the model selections for GNN which automatically builds a hierarchical ensemble model. 
AutoHEnsGNN uses proxy evaluation to evaluate many GNNs to generate a high performance GNN pool, then explore a hierarchical ensemble. 
However, since there is no exploration in graph data with heterophily on this method, its effectiveness for such graph data is unknown.\\

\noindent
Various GNNs based on GCN have been proposed to improve the accuracy of node classification in graph data. 
GAT\cite{gat} uses the attention mechanism to learn different weights for each neighborhood. 
GCN-LPA\cite{gcnlpa} uses Label Propagation for regularization. 
SGC\cite{sgc} reduces model complexity by removing non-linear transformations in GCN. 
Geom-GCN\cite{geom} uses the bi-level aggregation and extracts structural information of graph data using geometric relationships in latent space. 
GNN has a problem called over-smoothing that adversely affects accuracy when the number of layers of GNN is increased. 
To solve this problem, JKNet\cite{jknet} captures local information by combining representations of hidden layer.
DropEdge\cite{dropedge} employs a renormalized graph convolution matrix with randomly removed edges.
APPNP\cite{appnp} uses Personalized PageRank to capture information on a wide range of nodes without increasing the number of layers. 
GCNII\cite{gcnii} constructs a deep model with initial residual and identity mapping. 
As a GNN considering homophily and heterophily of graph data, CPGNN\cite{cpgnn} propagates pre-computed label predictions by a matrix that holds the probability that each label is connected. 
FAGCN\cite{fagcn} flexibly extracts information by combining high-pass filters and low-pass filters.\\

\noindent
Since there is little exploration in the problem about graph data with heterogeneity, GNN proposed in the early days may perform poorly on graph data with heterophily.
Besides, the traditional GNN may not be able to fully leverage the structural information and feature vector information that LP and MLP can capture. 
In HMSF, it can flexibly learn information of graph data by combining models that can address these problems.
Specifically, it basically selects H2GCN which can perform well on both graph data with homophily and heterophily and selects GCN only for graph data with high average degree that H2GCN performs poorly.
Furthermore, it selects CPF to leverage the structural information and feature vector information more fully than teacher GNN model, and do not select CPF only for graph data with heterophily that LP in CPF does not perform well.

\section{Experiments}\label{experiment}
In this section, we conduct experiments to prove the effectiveness of HMSF across multiple datasets. 
Specifically, we compare HMSF with traditional GNN by the accuracy of node classification in two dataset settings.

\subsection{Dataset}
We use the nine datasets summarized in Table \ref{tab:dataset}. 
Each dataset is frequently used as a benchmark in GNNs, and their details are as follows.

\begin{itemize}
\item \textbf{Cora, Citeseer, Pubmed (Citation Data)} \cite{pubmed, coracite} is a citation network of papers, where nodes represent papers, edges represent citation relations, and labels represent fields of papers. Bag-of-words is used as the feature vector.

\item \textbf{Texas, Wisconsin, Cornell} \cite{geom} is a network representing the link relationships of university web pages, where nodes represent web pages, edges represent link relations, and labels represent page categories. Bag-of-words is used as the feature vector.

\item \textbf{Squirrel, Chameleon} \cite{geom} is a network of Wikipedia link relations related to a specific topic, where web pages correspond to nodes, link relations to edges, and labels are classified from page traffic.

\item \textbf{Actor} \cite{geom} is a network of the co-occurrence of actors in Wikipedia, where actors correspond to nodes, page co-occurrences to edges, and labels are classified from words in Wikipedia pages.
\end{itemize}

\begin{table}[H]
    \begin{center}
\caption{Dataset Overview} \label{tab:dataset}
  \begin{tabular}{ c | c c c c } \hline
&Nodes&Edges&Classes&Number of Features\\ \hline
Cora&2708&5278&7&1433\\
Citeseer&3327&4552&6&3703\\
Pubmed&19717&44327&3&500\\
Texas&183&279&5&1703\\
Wisconsin&251&450&5&1703\\
Actor&4600&26659&5&932\\
Squirrel&5201&198353&5&2089\\
Chameleon&2277&31371&5&2325\\
Cornell&183&277&5&1703\\
\hline
\end{tabular}
\end{center}
\end{table}

\subsection{Experimental Setup}
We use two types of data splits: (i) 10 random splits dividing the training/test/validation data into 48\%/20\%/\\32\%, respectively, for each label based on H2GCN\cite{h2}, and (ii) the training data based on GCN\cite{gcn}, with 20 nodes for each label for training, 1000 for testing, and 500 for validation. 
For Texas, Wisconsin and Cornell which have few nodes, we use up to 5 nodes for each label for training, 100 for testing, and 50 for validation. 
For Cora, Citeseer and Pubmed, we use the same fixed data split as GCN. 
For other data, we create our own 10 different splits since GCN did not conduct experiments on these datasets. 
We run experiments on each dataset with 10 seeds and calculate the mean accuracy. 
We employ Adam\cite{adam} as an optimizer. 
We implement H2GCN  based on the model available on \url{https://github.com/GemsLab/H2GCN} and implement GCN and CPF from scratch. \\

\noindent
We use GCN with $K_{\mbox{\it gcn}} = $2 layers and H2GCN with $K_{\mbox{\it h2gcn}}\in\{1, 2\}$ layers, and set the dimension of the middle layer to $d_{\mbox{\it gnn}} = 64$ for both GNNs. 
We use the number of dropout rate $dr \in \{0.0, 0.5\}$, the weight decay $wd \in \{0.0005, 0.00001\}$, and the activation function of H2GCN $\sigma \in \{ $ReLU, None$\}$, and select the best combination of parameters based on the performance on validation dataset. 
We train the GNN for $e = 1000$ epochs and stop training if the validation loss does not decrease while 200 epochs.
We adopt the accuracy on the test data at the epoch when the validation accuracy is maximized. 
We use the CPF with a two-layer MLP, and set the dimension of the middle layer of the MLP to $d_{\mbox{\it mlp}} = 64$.
We use the number of dropout rate of the MLP $dr_{\mbox{\it mlp}} \in \{0.5, 0.8\}$, the number of layers of the PLP $K_2 = 8$, the dropout rate of the PLP $dr_{\mbox{\it plp}} = 0.8$, the learning rate $lr \in \{ 0.01, 0.001\}$, and the weight decay $wd \in \{0.01, 0.001\}$, and select the best combination of parameters based on the performance on validation dataset.
We train the CPF for $e = 2000$ epochs and stop training if the validation loss does not decrease while 200 epochs. 
We adopt the accuracy on the test data at the epoch when the CPF stops learning.
We also use $\beta \in \{2, 10, 50, 100\}$ and $\gamma \in \{0.2, 0.4, 0.6, 0.8\}$, and select the best combination as well.

\subsection{Experimental Results on the Splits Provided by H2GCN \cite{h2}} \label{sec:ef}

Table \ref{tab:result1} shows the values of indicators and the accuracies of GNN models incorporated in HMSF. 
We use the result of HMSF when $\beta = 10$ and $\gamma = 0.6$. 
The best results are highlighted in bold for each dataset.
It shows that Squirrel and Chameleon have a high average degree, and GCN is the best model on these datasets.
On the other hand, H2GCN performs well compared to GCN on other data which have low average degree.
CPF shows a lower accuracy than its teacher model on datasets where $h$ is less than 0.5, and CPF and its teacher model take almost close accuracy on datasets with $h$ greater than 0.5.
This is because this data split has many training data, and the teacher model can perform well.
Overall, we can observe that HMSF selects models that achieve high performance for each dataset.

\begin{table}[H]
   \caption{Experimental results of HMSF on the splits provided by H2GCN \cite{h2}. The parentheses in CPF indicate the GNN model used as the teacher model and H2 in Selected Model means H2GCN.
}
\label{tab:result1}
  \hspace{-1.2cm}
  \begin{tabular}{l | c c c c c c c c c } \hline
&texas&wisconsin&actor&squirrel&chameleon&cornell&Cora&citeseer&Pubmed\\ \hline
Average Degree&3.14&3.65&7.02&76.30&27.58&3.04&3.90&2.77&4.50\\
Actual Value $h$&0.06&0.18&0.22&0.22&0.23&0.30&0.81&0.74&0.80\\
Estimated Value $h'$&0.08&0.16&0.23&0.32&0.39&0.37&0.86&0.87&0.85\\ \hline\hline
HMSF& \textbf{84.86}& \textbf{85.29}& \textbf{35.20}& \textbf{49.26}& \textbf{66.38}& \textbf{81.62}&87.63&\textbf{76.76}&88.92\\
Selected Model&H2&H2&H2&GCN&GCN&H2&CPF(H2)&CPF(H2)&CPF(H2)\\ \hline
GCN&59.19&58.04&29.86&\textbf{49.26}&\textbf{66.38}&58.11& 87.14 &75.24 &87.84\\
CPF(GCN)&56.49&59.61&29.68&47.66&61.95&55.41&86.76&74.73&88.13\\
H2GCN&\textbf{84.86} &\textbf{85.29} &\textbf{35.20} &35.24 &57.21 &\textbf{81.62} &\textbf{87.89} &76.65 &\textbf{89.39}\\
CPF(H2GCN)&81.35&82.16&34.63&34.44&52.26&79.46&87.63&\textbf{76.76}&88.92\\
\hline
 \end{tabular}
\end{table}
\noindent
Table \ref{tab:result1comp} shows the comparison of accuracy between HMSF and other baseline methods for each dataset and mean accuracy.
Accuracy for models marked with `*' are the results from \cite{h2, gcnii, geom}.
HMSF has the 2.13\% higher mean accuracy than H2GCN-2 which has the highest mean accuracy among each method.
We can observe that the model which performs well for each dataset is different, and GCNII and Geom-GCN not incorporated in HMSF achieve the best performance on Cora, Citeseer and Pubmed.
However, the mean accuracy of HMSF is about 10\% higher than Geom-GCN.
Furthermore, the mean accuracy excluding Actor and Squirrel of HMSF increases 7.19\% compared to GCNII.
This result proves that HMSF can achieve stable performance by selecting the useful model for each dataset.

\begin{table}[H]
   \caption{Comparison of accuracy between HMSF and other baseline methods. * denotes the experimental results from \cite{h2, gcnii, geom}.
}
\label{tab:result1comp}
  \hspace{-0.5cm}
  \begin{tabular}{ l | c  c  c  c  c  c  c  c  c | c} \hline
&Texa.&Wisc.&Acto.&Squi.&Cham.&Corn.&Cora&Cite.&Pubm.&Mean Acc\\ \hline
HMSF& \textbf{84.86}&85.29& 35.20& \textbf{49.26}& \textbf{66.38}& 81.62&  87.89 &76.65& 89.39&\textbf{72.95} \\
GCN &59.19& 58.04& 29.86& \textbf{49.26}& \textbf{66.38}& 58.11& 87.14& 75.24& 87.84& 63.45\\
CPF(GCN)&56.49 &59.61& 29.68 &47.66 &61.95 &55.41& 86.76& 74.73& 88.13 &62.27\\
H2GCN &\textbf{84.86}& \textbf{85.29}& 35.20& 35.24& 57.21& \textbf{81.62}& 87.89 &76.65 &\textbf{89.39}& 70.37\\
CPF(H2GCN)&81.35& 82.16& 34.63& 34.44 &52.26& 79.46 &87.63 &\textbf{76.76} &88.92 &68.62\\ \hline
GAT*&58.38& 55.29& 26.28& 30.62& 54.69& 58.92&82.68& 75.46& 84.68&58.56\\
H2GCN-1*&\textbf{84.86}&\textbf{86.67}&\textbf{35.86}&36.42&57.11&\textbf{82.16}&86.92&77.07&89.40&70.72\\
H2GCN-2*&82.16&85.88&35.62&37.90&59.39&\textbf{82.16}&87.81&76.88&89.59&70.82\\
GCNII*&69.46&74.12&-&-&60.61&62.48&\textbf{88.49}&77.08&89.57&-\\
Geom-GCN-I*&57.58&58.24&29.09&33.32&60.31&56.76&85.19&\textbf{77.99}&\textbf{90.05}&60.95\\
Geom-GCN-P*&67.57&64.12&31.63&38.14&60.90&60.81&84.93&75.14&88.09&63.48\\ \hline

 \end{tabular}
\end{table}

\subsection{Experimental Results on the Splits Provided by GCN \cite{gcn}} \label{sec:few}

Table \ref{tab:result2} shows the values of indicators and the comparison of accuracy between HMSF and other baseline methods for each dataset and mean accuracy.
We use the result of HMSF when $\beta = 10$ and $\gamma = 0.6$. 
The best results are highlighted in bold for each dataset.
We summarize below what we observe from these results:

\begin{itemize}
\item The estimated value of the edge homophily ratio $h'$ is very close to the actual value $h$ for all datasets, and the error is at most 0.24 in Chameleon. 
This is sufficient as a predicted value to determine whether to select CPF.

\item On Cora, Citeseer, and Pubmed which have a value of $h'$ greater than 0.6, CPF has the highest accuracy. 
On the other hand, for other data which have a value of $h'$ less than 0.6 except Actor, Squirrel, and Cornell, we can observe that the accuracy of CPF is lower than its teacher model.
Besides, the accuracy of CPF is higher than its teacher model on Actor, Squirrel, and Cornell. 
This is because the teacher model cannot learn sufficiently since there is few training data on this data split, and the additional learning by MLP of CPF is more effective than the adverse effects of LP.

\item On Texas, Wisconsin, Actor, and Cornell, which have low $h'$ and low average degree, H2GCN shows higher accuracy compared to GCN.
On the other hand, GCN has a higher accuracy compared to H2GCN on Chameleon which has high average degree. 
Squirrel also has a high average degree, but the accuracies of GCN and H2GCN take close value.
This is because the training data on this data split setting is very few and the learning is not stable, so that the effective model on Squirrel changes depending on the data split.

\item The above observation proves that HMSF has the highest mean accuracy and can achieve stable performance by selecting the optimal model for each dataset.

\end{itemize}

\begin{table}[H]
    \caption{Experimental results of HMSF and other baseline methods on the splits provided by GCN \cite{gcn}. The parentheses in CPF indicate the GNN model used as the teacher model and H2 in Selected Model means H2GCN. 
}
\label{tab:result2}
  \hspace{-1.5cm}
  \begin{tabular}{ l | c  c  c  c  c  c c  c  c | c } \hline
&Texa.&Wisc.&Acto.&Squi.&Cham.&Corn.&Cora&Cite.&Pubm.& Mean Acc\\ \hline
Average Degree&3.14&3.65&7.02&76.30&27.58&3.04&3.90&2.77&4.50&-\\
Actual Value $h$&0.06&0.18&0.22&0.22&0.23&0.30&0.81&0.74&0.80&-\\
Estimated Value $h'$&0.12&0.23&0.34&0.39&0.47&0.27&0.83&0.82&0.87&-\\
\hline \hline
HMSF&\textbf{78.10}&\textbf{74.00}&25.03&28.19&\textbf{50.71}&67.70&\textbf{83.51}&72.17&\textbf{80.57}&\textbf{62.22}\\
Selected Model&H2&H2&H2&GCN&GCN&H2&CPF(H2)&CPF(H2)&CPF(H2)&-\\\hline
GCN&53.50&43.80&23.94&28.19&\textbf{50.71}&55.20& 81.82& 71.80& 79.19&54.24\\
CPF(GCN)&54.40&43.70&24.58&27.01&46.43&57.20&82.68& \textbf{72.59}& 80.23 &54.31\\
H2GCN&\textbf{78.10}&\textbf{74.00}&25.03&28.65&44.55&67.70&80.87& 70.76& 79.84 &61.06\\
CPF(H2GCN)&77.10&70.30&\textbf{25.15}&\textbf{30.60}&41.84&\textbf{72.50}&\textbf{83.51}& 72.17& \textbf{80.57}&61.53\\
\hline
 \end{tabular}
\end{table}

\subsection{Analysis of Feature Thinning at Nodes with High Average Degree}
\begin{figure}[H]
\begin{center}
\includegraphics[width=10.0cm]{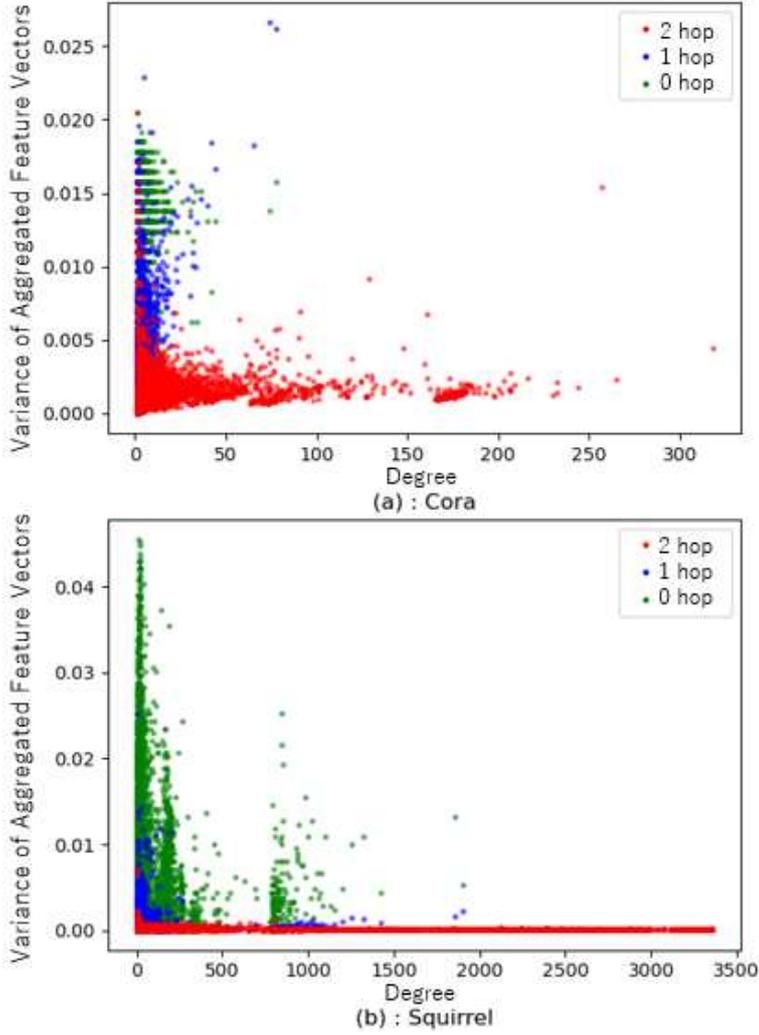}
  \caption{Relationship between degree and variance}\label{fig:bunsan}
\end{center}
\end{figure}

We investigate whether the features at nodes with high degree are thinned when aggregating the features of neighboring nodes.
We illustrate the relationship between degree and variance in Figure \ref{fig:bunsan} by calculating the variance of the feature vector which aggregates the features of the 0 hop (self-loop), 1 hop, and 2 hop neighboring nodes for all nodes on the graph.
We used Cora (in Figure \ref{fig:bunsan}(a)) and Squirrel (in Figure \ref{fig:bunsan}(b)) based on the data splits provided by H2GCN \cite{h2}, and set the horizontal axis as the degree of each node and the vertical axis as the variance of the feature vector. 
Note that points of 0 hop and 1 hop  use the degree of 1 hop away, whereas points of 2 hop use the degree of 2 hop away.

\noindent
On both datasets, the variances of 0 hop are high, and the variances of 1 hop tend to have low value compared to 0hop.
Furthermore, in the case of 2 hop, the degree is much larger than 1 hop, and the variance is lower than 1 hop. 
Comparing (a) and (b), there are many 1 hop nodes which have variance exceeding 0.01 in (a), whereas most nodes have variance below 0.01 in (b).
In addition, regarding 2 hop, most nodes in (a) have degree between 0 and 200 and variance between 0.0 and 0.005; in (b), on the contrary, there are many nodes with degree exceeding 200 and variance close to 0.0.\\

\noindent
As observed from a result of above, the variances of the feature vector become low if nodes have a high degree.
The feature vector with low variance means that the amount of features of vector is flattened, and H2GCN becomes difficult to distinguish the difference between features of aggregated vectors compared to 0 hop vectors.
Therefore, we prove that H2GCN is far from the optimal model for graph data with high average degree.

\subsection{Analysis of Label Propagation Suitability}
\begin{figure}[H]
\begin{center}
\includegraphics[width = 10.0cm]{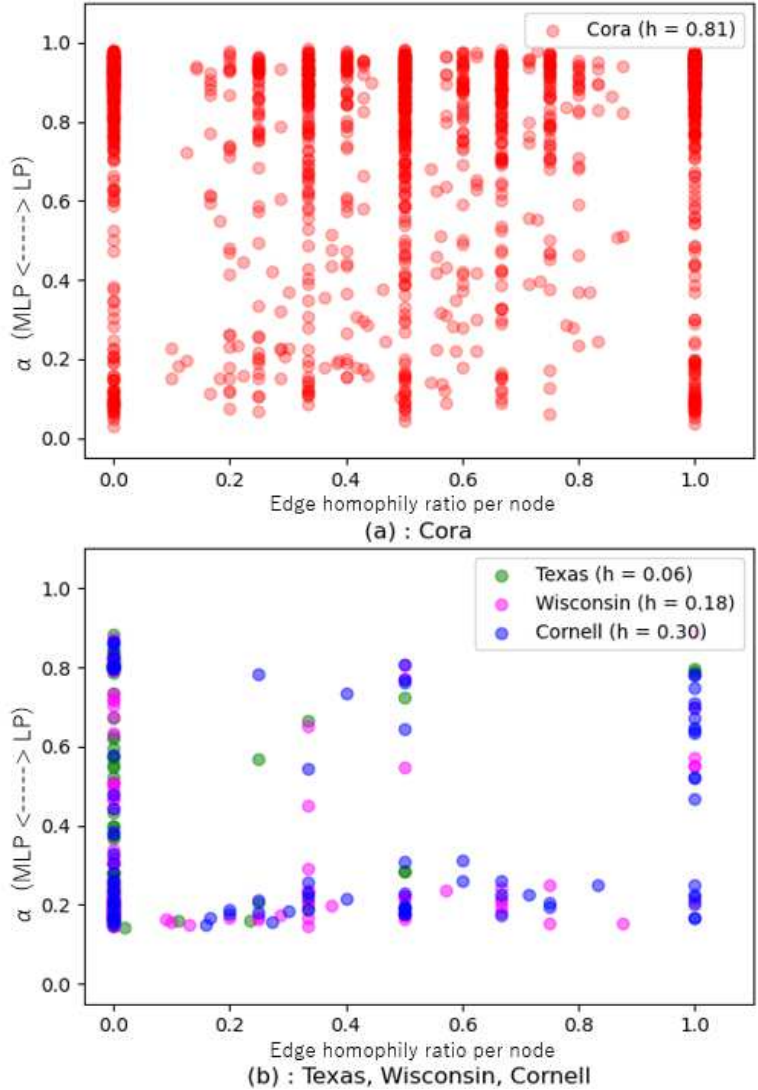}
  \caption{Relationship between homophily ratio at node level and alpha value}\label{fig:cpf}
\end{center}
\end{figure}

We investigate the reason why CPF perform poorly compared to its teacher model on graph data with a low edge homophily ratio.
Figure \ref{fig:cpf} illustrates the relationship between the edge homophily ratio per node and $\alpha$ which is a learnable balance parameter of trained CPF for each node.
We use Cora (in Figure \ref{fig:cpf}(a)) as a dataset with a high edge homophily ratio, and Texas, Wisconsin, Cornell (in Figure \ref{fig:cpf}(b)) as a dataset with a low edge homophily ratio on the data splits provided by H2GCN \cite{h2}.
The edge homophily ratio per node for node $x$ indicates the ratio of edges connecting to a node which has the same label among edges connected to a node $x$.
The parameter $\alpha\in [0, 1]$ in CPF is a balance parameter that adjusts the output ratio of MLP and LP for each node.
Specifically, $\alpha$ close to 1.0 means that LP is more effective than MLP, and $\alpha$ close to 0.0 means that MLP is more effective than LP on CPF. 

\noindent
Comparing (a) and (b) in Figure \ref{fig:cpf}, many nodes assign large weights on LP in (a), many nodes assign large weights on MLP in (b).
This result indicates that LP is not  effective for graph data with a low edge homophily ratio compared to MLP.
In the case such as (b), CPF uses almost MLP to train the student model. 
However, MLP is less expressive than teacher GNN model, so it is difficult for CPF to achieve better performance than its teacher model. 
Furthermore, especially in H2GCN which perform well on graph data like (b), it already has a mechanism like MLP, there is almost no information from feature vectors that can be captured only by MLP.
These results prove that selecting CPF for graph data with a low edge homophily ratio does not improve the result, but rather likely to decreases compared to its teacher model.

\section{Conclusion}
In this paper, we have proposed a hierarchical model selection framework (HMSF) by analyzing the effectiveness of GNN models based on average degree and edge homophily ratio.
We have demonstrated that HMSF can select an appropriate model and achieve the highest mean accuracy in evaluation experiments across multiple benchmark datasets.
We have also proved that the variance of the feature vector which aggregates the features of neighboring nodes takes small value on nodes with high degree, and LP in CPF is not effective on graph data with low edge homophily ratio.\\

\noindent
In the experiment on the setting of data splits with low training data, GCN or H2GCN is selected for Actor, Squirrel, and Cornell in our HMSF, but CPF(H2GCN) outperforms its teacher model H2GCN even on graph data with heterophily.
As a future work, we consider another indicator to select the appropriate CPF model for low training data.
In addition, we need to improve the effectiveness of HMSF by incorporating new GNN models which achieve state-of-the-art performances.

\small
\bibliographystyle{junsrt}
\bibliography{firstinitial_gakkai}

\end{document}